\documentclass[runningheads]{llncs}
\usepackage[T1]{fontenc}
\usepackage{graphicx}
\usepackage{url}
\usepackage{hyperref}
\usepackage{cite}
\usepackage{amsmath,amssymb,amsfonts}
\usepackage{algorithmic}
\usepackage{textcomp}
\usepackage{xcolor}
\usepackage{booktabs}
\usepackage{multirow}
\begin{document}
\title{Trustworthy Pedestrian Trajectory Prediction via Pattern-Aware Interaction Modeling}
\titlerunning{InSyn}
\author{
Kaiyuan Zhai\inst{1,2} \and
Juan Chen\inst{1}\thanks{Corresponding Author} \and
Chao Wang\inst{1}$^{\star}$\and
Zeyi Xu\inst{1} \and
Guoming Tang\inst{2}
}
\authorrunning{Zhai et al.}
\institute{Shanghai University \and
The Hong Kong University of Science and Technology (Guangzhou)}
\maketitle              % typeset the header of the contribution
\begin{abstract}
Accurate and reliable pedestrian trajectory prediction is critical for the application of intelligent applications, yet achieving trustworthy prediction remains highly challenging due to the complexity of interactions among pedestrians. Previous methods often adopt black-box modeling of pedestrian interactions. Despite their strong performance, such opaque modeling limits the reliability of predictions in real-world deployments. To address this issue, we propose InSyn (Interaction-Synchronization Network), a novel Transformer-based model that explicitly captures diverse interaction patterns (e.g., walking in sync or conflicting) while effectively modeling direction-sensitive social behaviors. Additionally, we introduce a training strategy, termed Seq-Start of Seq (SSOS), designed to alleviate the common issue of initial-step divergence in numerical time-series prediction. Experiments on the ETH and UCY datasets demonstrate that our model not only outperforms recent black-box baselines in prediction accuracy, especially under high-density scenarios, but also provides transparent interaction modeling, as shown in the case study. Furthermore, the SSOS strategy proves to be effective in improving sequential prediction performance, reducing the initial-step prediction error by approximately 6.58\%. Code is avaliable at \url{https://github.com/rickzky1001/InSyn}

\end{abstract}
\section{Introduction}

Pedestrian trajectory prediction is essential for real-world applications such as autonomous driving \cite{luo2018porca} and robotic navigation \cite{bae2023sit}. Accurate and reliable forecasting enables intelligent systems to understand human behavior better and ensures trustworthy predictions. However, the presence of complex pedestrian interactions poses significant challenges to the task. In a given scene, pedestrians may respond to nearby individuals in different ways~\cite{korbmacher2022review}. For instance, the trajectories of two pedestrians may exhibit conflict. Alternatively, they may show weak or no influence on each other, such as when walking in sync or as part of a group. In particular, even in walk-in-sync scenarios, subtle interactions may still occur. Effectively modeling these intricate interaction patterns, especially in high-density environments, remains a major challenge in pedestrian trajectory prediction.
\vspace{-8pt}
\begin{figure}[h]
  \centering
  \includegraphics[width=4.7in]{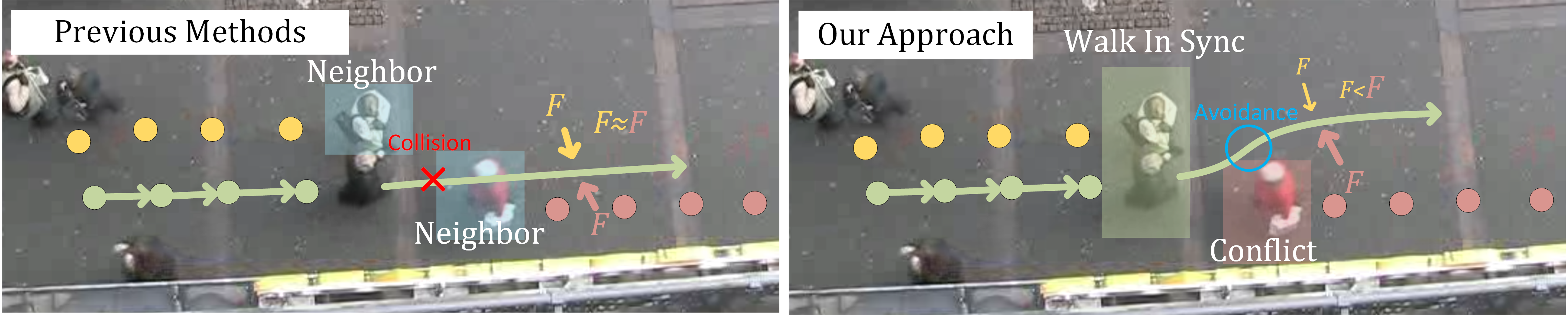}
  \caption{
    Comparison of Interaction Modeling: \textbf{Previous Methods vs. Our Approach.} \textit{F} represents the interaction effect between pedestrians. In traditional approaches (top), all neighbors of the agent are treated as being in the same state. Our method (bottom) introduces a more refined modeling strategy by considering the specific states of neighboring pedestrians, providing a more nuanced understanding of the neighboring interaction.
  }
  \vspace{-18pt}
  \label{intro}
\end{figure}

Recently, extensive studies have investigated pedestrian interactions \cite{alahi2016social,wang2023trajectory,sang2024mstcnn,moon2024should,yue2022human}, using approaches like social pooling layers \cite{alahi2016social,daniel2021pecnet}, social force mechanism \cite{yue2022human,zhang2023forceformer}, and graph neural networks (GNNs) \cite{yu2020spatio,lin2025multi,benrachou2025graph}. However, a common characteristic of these methods is that they rely on features like relative position between an agent and its neighbors to model influence, treat all neighboring pedestrians by black-box representation, failing to differentiate patterns such as \textit{In Sync} and \textit{Conflict} (see Figure~\ref{intro}). This oversimplification may lead to overfitting and untrustworthy predictions. For example, failing to recognize synchronized walking (e.g., friends walking side-by-side) and mistakenly modeling them with repulsive forces can lead to unrealistic divergence in their predicted paths. This black-box modeling not only reduces prediction accuracy, but also limits interpretability and trustworthiness in real-world scenarios. These limitations become particularly pronounced in crowded or dynamic environments.

To address this issue, we need a method that \textit{understands what’s exactly going on between pedestrians}—recognizing interaction patterns rather than treating them as abstract black-box inputs—to ensure more reliable and accurate predictions (see Figure~\ref{intro}).

Although pedestrian trajectories are highly stochastic, they usually have a clear goal \cite{tomasello2005understanding}. Focusing on the target goal can significantly enhance model performance. Building on this insight and the suitability of the attention mechanism for capturing long-range dependencies in sequential data \cite{giuliari2021transformer}, we propose \textbf{InSyn}, a Transformer-based model.

The model consists of three components: (1) \textbf{Interaction Encoder}: designed to explicitly extract interaction information and integrates interaction features, goal-driven behavior, and observed trajectories through the self-attention mechanism. (2) \textbf{Trajectory Generator}: forms an cross-attention mechanism in conjunction with the Interaction Encoder, and incorporates the proposed SSOS strategy to alleviate divergence at the initial prediction step. (3) \textbf{Seq-CVAE Goal Sampler}: a conditional generative model specifically designed for sequential prediction, used for goal sampling.

% We evaluate our model on the well-established pedestrian trajectory prediction datasets ETH \cite{pellegrini2009you} and UCY \cite{lerner2007crowds}. The experimental results demonstrate that our model outperforms recent methods in key metrics, particularly in scenarios with complex interactions. We further conduct ablation studies to show the superiority of interaction modeling and the SSOS strategy. Additionally, the case study highlights the interpretability of our approach by illustrating how specific interaction patterns influence prediction outcomes.

In summary, our work makes two key contributions:
\begin{itemize}
  \item We propose a pedestrian interaction modeling approach that explicitly identifies and leverages specific interaction patterns, achieving a significant improvement in average ADE compared to the previous black-box baselines and contributing to more trustworthy socially-aware trajectory prediction.
  \item For numerical time-series prediction tasks, we introduce a novel training strategy for the Transformer encoder-decoder architecture. This strategy, termed SSOS, mitigates the divergence in the first prediction step, thereby reducing error accumulation and improving overall performance.
\end{itemize}

\section{Related Work}

Pedestrian interaction is a key factor influencing prediction accuracy. Pedestrian motion is not only driven by individual goals but also affected by interactions with surrounding pedestrians.

Current mainstream methods for interaction modeling include social pooling layers \cite{alahi2016social,yang2020tppo,hsu2025pedestrian}, GNNs \cite{lin2025multi,yu2020spatio,sang2025review,wang2025hsigcn}, and attention mechanisms \cite{li2021spatio,yang2024meta,pang2025data}. Among these, the social attention module aggregates interaction information by analyzing correlations between pedestrian motion and future trajectories \cite{yang2020tppo}. Yang et al. \cite{yang2024meta} introduced a social graph attention mechanism combined with a pseudo-oracle predictor to capture social interactions and intention states, enhancing prediction accuracy. Transformers have been utilized in pedestrian interaction modeling \cite{wang2024multimodal,yang2023long,jiang2024social,amirloo2022latentformer,su2024unified}. For example, Yuan et al. \cite{yuan2021agentformer} proposed AgentFormer that learns spatiotemporal interaction embeddings from sequential trajectory features. Similarly, Yang et al. \cite{yang2023long} leveraged GNNs and Transformers to model spatial and temporal dependencies. To address future interaction modeling, Amirloo et al. \cite{amirloo2022latentformer} employed the self-attention mechanism of Transformer to model pedestrian interactions and considered future states autoregressively during decoding to avoid trajectory conflicts. 

However, most of these methods rely on black-box representations of social influence. This limits the model’s ability to produce reliable and interpretable predictions. In contrast, our approach explicitly leverages structured interaction patterns, such as walking \textit{In Sync} or engaging in \textit{Conflict}, aiming to enhance the trustworthiness and transparency of socially-aware trajectory prediction.

\section{Problem Formulation}

\begin{figure}[h]
  \centering
  \includegraphics[width=4.7in]{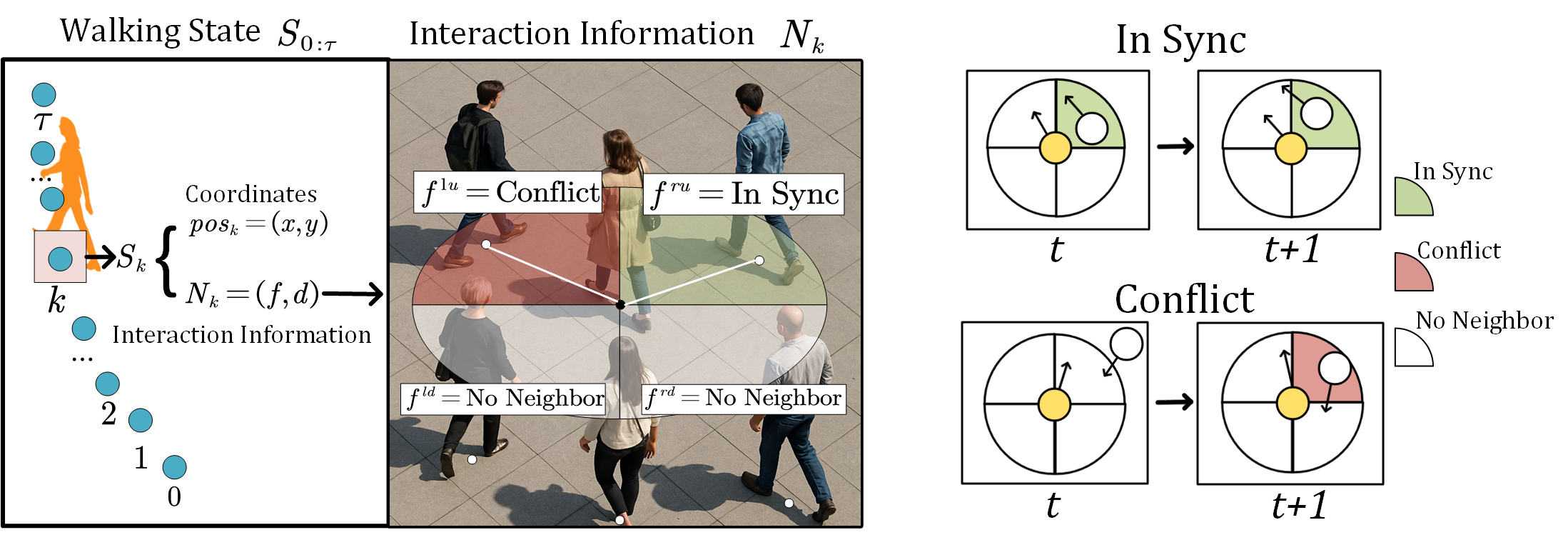}
  \caption{\textbf{Left} is the illustration of the input $S_{0:\tau}$. At each time step, the walking state $S_k$ comprises the 2D coordinates $(x, y)$ and interaction information $N_k$. \textbf{Right} demonstrates the scenarios of \textit{In Sync}, \textit{Conflict} and \textit{No Neighbor} state. Note that the 4-regions partition and the scenarios identification use simple spatial-temporal rules for transparency, more complex methods could be applied here but beyond our scope.}
  \vspace{-15pt}
  \label{problem_formulation}
\end{figure}

\label{Problem Formulation}
In this study, pedestrian trajectory prediction is formulated as follows: given the walking states $S_{0:\tau}$ of a certain agent during the observation time steps $\mathord{0:\tau}$, the model outputs the predicted trajectory $\widehat{pos}_{\tau+1:T}$ of the agent for the future time steps $\mathord{\tau+1:T}$. The input walking states of agent $p$ are denoted as  $\mathcal{S}^p = \left( S_{0}^{p}, S_{1}^{p}, \dots, S_{\tau}^{p} \right)$, where $S_{t}^{p}$ represents the walking state at time step $t\ (0 \leq t \leq \tau)$. Each walking state $S_{t}^{p}$ consists of the agent's position $pos_t = (x_t, y_t)$ in 2D space and the pedestrian interaction information $N_t$. Since interactions are directional, we adopt a region partition strategy~\cite{alahi2016social} around each agent. The interaction information $N_t$ is defined as  
\begin{equation}
N_t = \left\{ (f_t^r, d_t^r) \right\}_{r \in \{\mathrm{lu, ru, ld, rd}\}}
\end{equation}

where $f_{t}^{r}$ represents the interaction state in region $r$ at time $t$; $d_{t}^{r}$ indicates the distances to the nearest neighbor; $\{\mathrm{lu, ru, ld, rd}\}$ represent the left-up, right-up, left-down, right-down of the agent respectively. 

The interaction state $f_{t}^{r}$ can be \textit{No Neighbor}, \textit{In Sync}, and \textit{Conflict}, as shown in Figure~\ref{problem_formulation}. The construction of interaction patterns is highly transparent. Specifically, if the nearest neighbor in a given region at time $t$ is the same as the one at time $t-1$, the interaction state is classified as \textit{In Sync}; otherwise, it is marked as \textit{Conflict}. A region with no pedestrians is classified as \textit{No Neighbor}, and a large distance value is assigned to indicate minimal influence. In cases of \textit{No Neighbor} or \textit{In Sync}, the interaction influence of the neighbor on the agent’s future trajectory may be limited. However, if a new pedestrian suddenly enters a region, referred to as \textit{Conflict}, it is more likely to impact the agent's trajectory. This identification uses simple rules for transparency and it could be replaced by a model-based method.

The goal of this study is to train a generative model
$
m_{\theta}\left( \widehat{pos}_{\tau+1:T} \mid S_{0:\tau} \right)
$, aiming to predict the distribution of pedestrians' future trajectory over time steps $\mathord{\tau+1:T}$, conditioned on the walking states during the observation period $\mathord{0:\tau}$.

\section{Methodology}

\begin{figure*}[h]
  % \vspace{0.2in}
  \centering
  \includegraphics[width=4.7in]{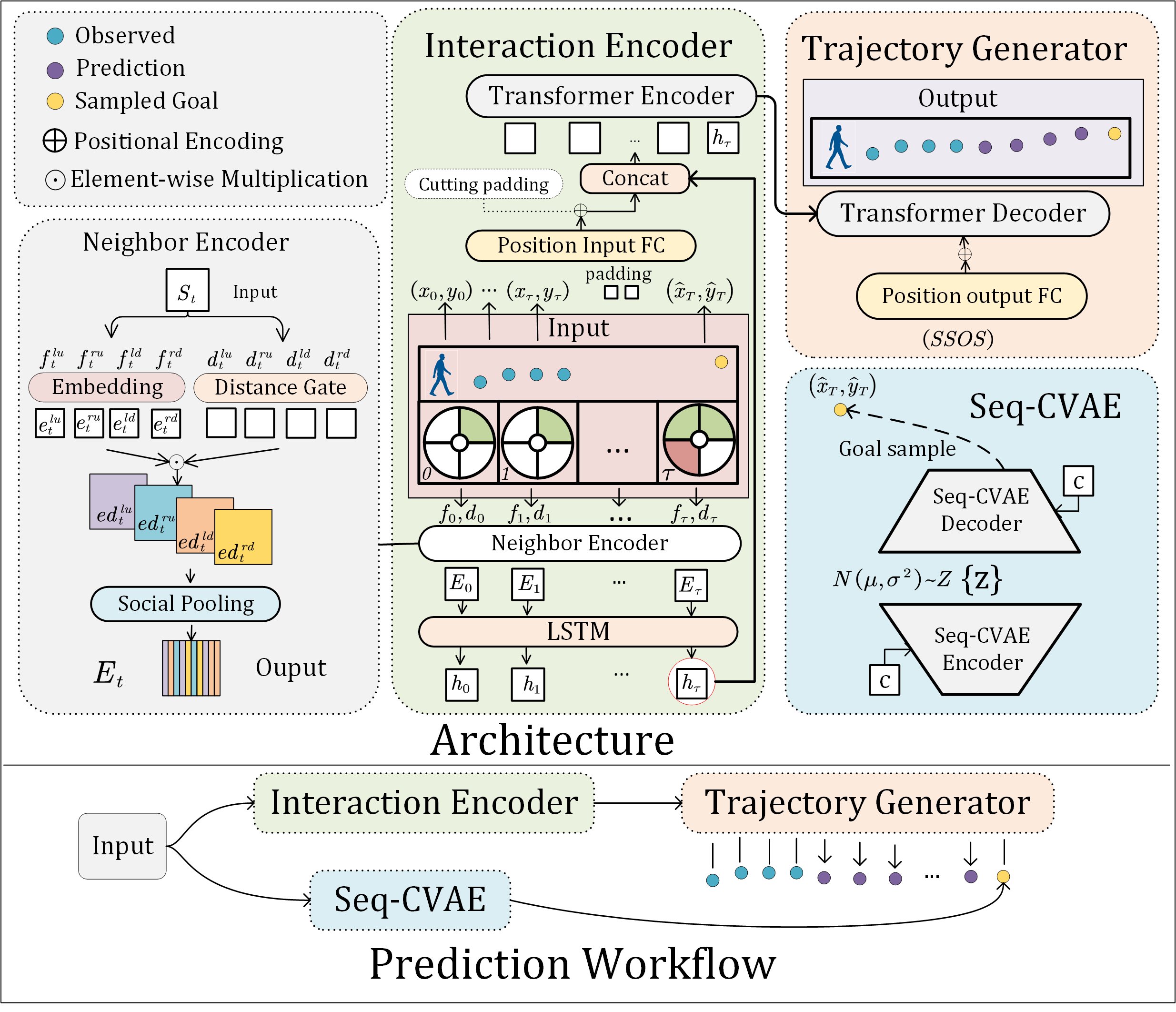}
  \caption{
    Overview of the \textbf{InSyn framework} for trajectory prediction. Our model consists of three key modules: (1) Interaction Encoder, (2) Trajectory Generator, and (3) Seq-CVAE. The input observed walking state includes the agent's trajectory positions $pos_{0:\tau}$ and its interaction information $ N_{0:\tau}$ within the observed time $0:\tau$.
  }
  \vspace{-15pt}
  \label{framework}
\end{figure*}
% \subsection{Framework Design}

In this section, we propose \textbf{InSyn}, a Transformer-based model tailored for the interpretable modeling of specific social interactions. The model consists of three components: (1) \textbf{Interaction Encoder}: Extracts interaction features and enables goal-driven prediction. (2) \textbf{Trajectory Generator}: Employing the SSOS strategy to mitigate initial prediction step divergence. (3) \textbf{Seq-CVAE Goal Sampler}: A generator for sequential data that utilizes high-dimensional latent variables for goal sampling. The framework of InSyn is illustrated in Figure~\ref{framework}, and the Seq-CVAE is depicted in Figure~\ref{seqCVAE}.

% We first discuss how our model captures specific interactions and achieves goal-driven functionality. We then elaborate on the motivation behind the SSOS strategy and, finally, the design of the Seq-CVAE module.

\subsection{Interaction Encoder}

The Interaction Encoder integrates observed trajectory, interaction information, and sampled goal to extract spatiotemporal features. These features are utilized for goal-driven and interaction-sensitive trajectory prediction. 

\subsubsection{Goal-Driven.} 

Given the history position information $pos_{0:\tau}^{p}$, the goal $\widehat{pos}_T$ generated by Seq-CVAE (introduced in Section ~\ref{sec:seq-CVAE}) is combined with the position information through concatenation. Notably, since self-attention cannot inherently capture positional information, positional encoding is used to encode the input positions. This encoding is based on the sequence positions of the inputs. Therefore, to ensure the sampled goal $\widehat{pos}_T$ is assigned the correct positional encoding, a $T-\tau-1$ length padding tokens must be inserted between the observed trajectory $pos{0:\tau}^{p}$ and the goal before concatenation.

\subsubsection{Interaction.} 
To facilitate trustworthy modeling of social interactions, we introduce a modular neighbor encoder that encodes pattern-aware effects.

The interaction patterns between pedestrians are extracted through a neighbor encoder module followed by an LSTM. The agent $p$'s observed social interaction information, represented as $\mathcal{N}^p = \left(N_{0}^{p}, N_{1}^{p}, ..., N_{\tau}^{p} \right) $, are encoded using the neighbor encoder. which consists of three components: an embedding layer, a Distance Gate, and a Social Pooling Layer, as shown in Figure~\ref{framework}. The embedding layer maps interaction states $f_{t}^{r}$ to learnable representations. These representations are then combined with the processed distance $d$ through element-wise multiplication. The process of $d$ is through the Distance Gate: 
$\mathrm{Gate}(d) = \sigma (\mathbf{W} \cdot d + \mathbf{b})$, where $\sigma(z) = \frac{1}{1 + e^{-z}}$ and $\mathbf{W},\mathbf{b}$ represent the parameters of linear layer. Inspired by the gating mechanism in LSTM~\cite{hochreiter1997long}, the Distance Gate maps $d$ to a weight in [0,1], reflecting distance-based interaction intensity.

After forming four regional interaction representations, they are passed through the Social Pooling Layer, which applies a max-pooling function across the region dimension. This enables the model to focus on the most influential interactions while suppressing less relevant or redundant information. The complete architecture of the Neighbor Encoder is illustrated in Figure \ref{framework}.

The social interaction information $N_{0}^{p}, N_{1}^{p}, ..., N_{\tau}^{p}$ for agent $p$ is processed by the Neighbor Encoder to produce the output $E_{0}^{p}, E_{1}^{p}, ..., E_{\tau}^{p}$, which is then fed into the LSTM. The hidden state of LSTM at the final observed time step $h_\tau$ contains all temporal information, thus no need for positional encoding. Consequently, $h_\tau$ is directly concatenated with the positionally encoded representation (see Figure~\ref{framework}).

\subsection{Trajectory Generator}
\label{sec:TG}
The Trajectory Generator incorporates the SSOS Strategy into the Transformer Decoder \cite{vaswani2017attention} to alleviate initial prediction divergence.
The Transformer's decoder relies on the Start of Sequence (SOS) as the initial input when generating the first output. In Natural Language Processing (NLP), the Begin-of-Sentence token $\left< bos \right>$ is commonly used as the SOS, which aligns with the semantic structure of text. However, in sequential trajectory prediction, the selection of SOS should consider the spatial information; otherwise, it may introduce noise. Giuliari et al. \cite{giuliari2021transformer} set the SOS to $(0, 0)$ for trajectory prediction. While $(0, 0)$ represents a fixed position in spatial coordinates, using it as the SOS may introduce noise and mislead the model due to its lack of alignment with the observed trajectory. Therefore, it is reasonable to consider using the position at the last observed time step, $\widehat{pos}_{\tau}$, as the SOS token of the decoder.

Nevertheless, we note that when the Transformer's encoder-decoder architecture is applied to trajectory prediction, a numerical time-series prediction task, using a single value as the SOS can lead to a lack of smoothness in the transition between the first predicted value $\widehat{pos}_{\tau+1}$ and the observed trajectory $pos_{0:\tau}$. That is to say, $\widehat{pos}_{\tau+1}$ may deviate noticeably from the last observation point $pos_{\tau}$. This issue likely arises because, during the initial decoding stage, the Transformer decoder relies solely on the self-attention with the SOS and the cross-attention with the encoder's output. Insufficient information or excessive noise in the SOS may lead to alignment bias, thus reducing prediction accuracy.

% \begin{figure}[h]
%   \centering
%   \includegraphics[width=3.4in]{Figure/SSOS.png}
%   \caption{
%     Comparison of SSOS and SOS.
%   }
%   \label{SSOS}
% \end{figure}

To mitigate this issue, we propose using the sequence data $pos_{0:\tau}$ as the SOS. We refer to this approach as SSOS and use it during training. In addition to computing the loss of the predicted trajectory $\widehat{pos}_{\tau+1:T-1}$, the loss of the reconstructed observed trajectory $\widehat{pos}_{1:\tau}$ is also minimized through gradient descent. The loss function using SSOS is presented below:
\begin{align}
  \mathcal{L}_{\text{SSOS}} &= \lambda_1 \cdot \text{MSE}(\widehat{pos}_{1:\tau}, pos_{1:\tau}) \nonumber \\
  &\quad + \lambda_2 \cdot \text{MSE}(\widehat{pos}_{\tau+1:T-1}, pos_{\tau+1:T-1})\label{loss}
\end{align}
where 
$\widehat{pos}_{1:\tau}$ denotes the reconstructed observed trajectory from time $1$ to $\tau$; 
$\widehat{pos}_{\tau+1:T-1}$ denotes the predicted trajectory excluding the sampled goal; 
$\lambda_1$ and $\lambda_2$ are used for balancing.

Ablation results in Section \ref{sec:ablationstudy} shows that this strategy enchances the performance of the model. However, the SSOS strategy generates additional outputs $\widehat{pos}_{1:\tau}$, which are not part of the target prediction.  These byproducts may introduce additional computational overhead.

\subsection{Seq-CVAE Goal Sampler}
\label{sec:seq-CVAE}
\begin{figure}[h]
  \centering
  \includegraphics[width=2.8in]{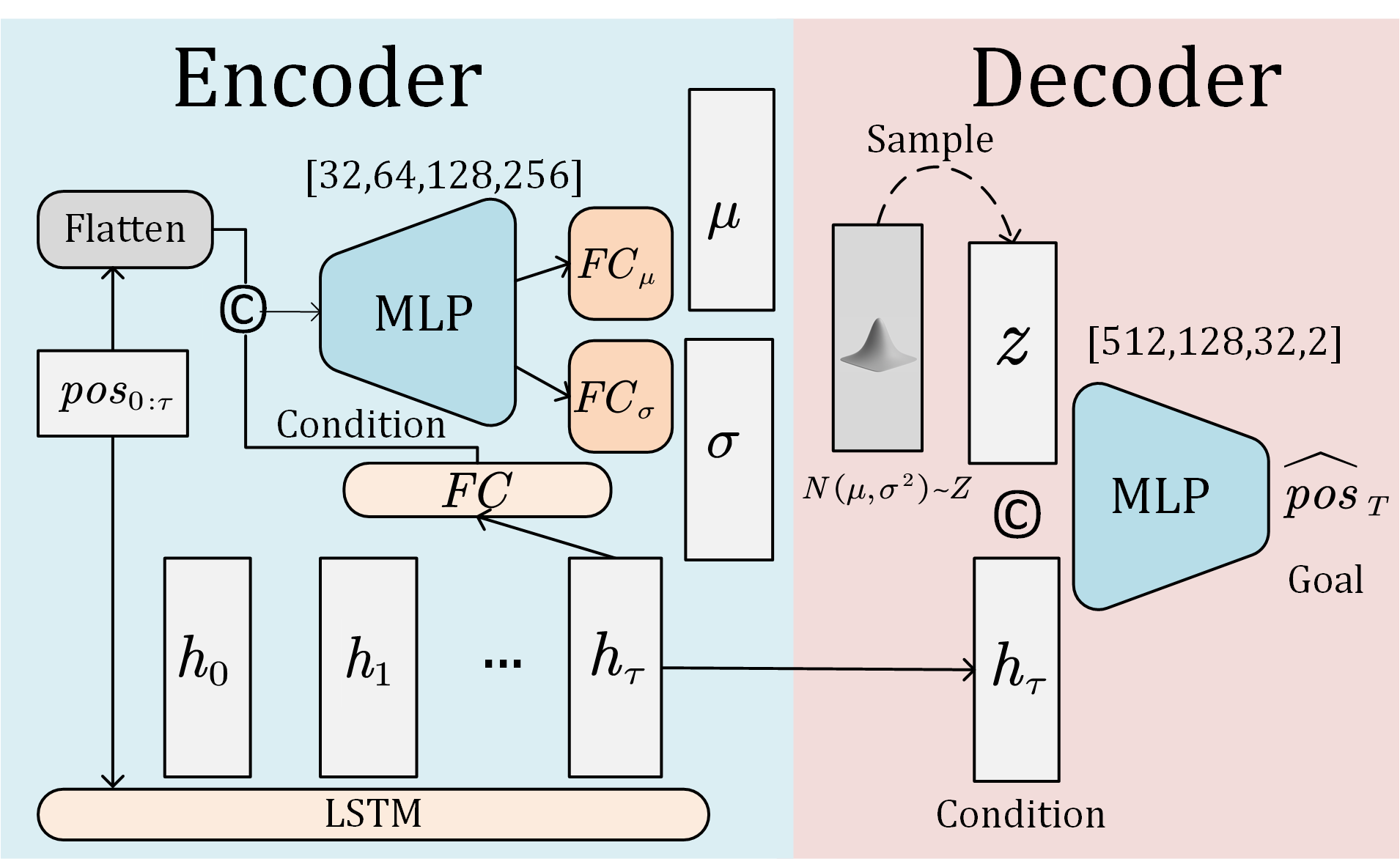}
  \caption{\textbf{Seq-CVAE Architecture}. Flatten represents flattening the input to a one-dimensional vector; MLP refers to the multi-layer perceptron, and $[a,b,c]$ above it indicates the dimensional transformations across its layers; $\copyright$ represents concatenation; $\mu$ and $\sigma$ represent the mean and standard deviation of the latent variable $z$. During training, the reparameterization trick \cite{kingma2013auto} is employed to enable backpropagation.}
  \label{seqCVAE}
  \vspace{-10pt}
\end{figure}

RNN (typically LSTM \cite{hochreiter1997long}) and Conditional Variational Autoencoder (CVAE) \cite{sohn2015learning,kingma2013auto} have been widely applied in pedestrian trajectory prediction \cite{yang2024ia,yue2022human,alahi2016social}. We combine the two approaches and propose a sequence-based generator, Seq-CVAE (Sequence CVAE), which is designed for sampling trajectory goals. The model predicts goals given the condition $c$, which is derived from the observed trajectory $pos_{0:\tau}$. These goals are subsequently fed into the Interaction Encoder to achieve its goal-driven functionality. Our purpose in this section is to learn a generative model $g_{\phi}(\widehat{pos}_T|c)$, where ${\phi}$ represents the model parameters. This model generates trajectory goals $\widehat{pos}_T$ that conform to the distribution of the conditional variable $c$. The Seq-CVAE architecture is illustrated in Figure~\ref{seqCVAE}.  

CVAE was first introduced in the field of image processing \cite{sohn2015learning}. It compresses high-dimensional data, such as pixel-based images, into a low-dimensional latent space while preserving key features of the image. However, unlike images, trajectories typically consist of low-dimensional data, such as the coordinates in $pos_{0:\tau}$. To extract latent features like velocity, direction, and acceleration, these inputs require dimensionality expansion rather than compression. Consequently, we propose modifying the CVAE Encoder's dimensionality reduction mechanism to perform dimensionality expansion instead.

Furthermore, regarding the construction of condition $c$ in CVAE, Yue et al. \cite{yue2022human} extract features from observed trajectory using an MLP. To better preserve the temporal feature of the trajectory, we adopt an LSTM module. To maintain information balance within the Encoder, the last step hidden state $h_\tau$ of the LSTM is passed through a fully connected layer for dimensionality reduction before concatenation, as shown in Figure~\ref{seqCVAE}. Therefore, for the Seq-CVAE, the condition $c$ for the encoder and decoder is defined as 
\begin{equation}
  c_{Encoder} = \mathrm{\textit{FC}}(\mathrm{LSTM}(h_\tau, \{pos_{0:\tau}\})), \quad
  c_{Decoder} = \mathrm{LSTM}(h_\tau, \{pos_{0:\tau}\})
\end{equation}
where $FC$ represents the fully connected layer, $h_\tau$ denotes the hidden state of LSTM at the last observed time step $\tau$, and $pos_{0:\tau}$ represents the coordinates of the observed trajectory.
\vspace{-10pt}
% \vspace{-15pt}
\section{Experiment}
% \vspace{-20pt}
% \subsubsection{Datasets.} We use the well-established datasets in the field of pedestrian trajectory prediction: ETH \cite{pellegrini2009you} and UCY \cite{lerner2007crowds}. The datasets include five sub-datasets, covering four different scenarios and 1536 trajectories. These datasets allow focused evaluation of pedestrian-pedestrian interactions without the confounding effects of complex scene elements. Similar to previous work \cite{wang2022stepwise,yue2022human,yuan2021agentformer,chib2025lg}, we adopt the leave-one-out strategy for splitting the training and testing sets. That is, four datasets are used for training and validation, and one dataset is used for testing. Since partitioning the neighboring regions requires real-world distances, we convert the coordinate scale to real-world metrics in meters. Due to the different frame rates of the ETH and UCY datasets, we unify both to 2.5 fps, corresponding to a time step of 0.4 seconds.

% \subsubsection{Evaluation Protocol.} Like many previous work \cite{yang2024ia,yu2020spatio,yuan2021agentformer}, visual information is not used in our approach. During the test phase, our model takes the trajectory within 3.2 seconds as input and outputs the trajectory for the subsequent 4.8 seconds. For fair comparison, we adopt the best-of-$K$ protocol, which is sampling $K = 20$ times and choosing the best result for metric computation. This is the standard protocol in pedestrian trajectory prediction \cite{li2019conditional,yue2022human,daniel2021pecnet,yu2020spatio}.

\subsubsection{Datasets and Evaluation Protocol.}
We evaluate on the standard ETH \cite{pellegrini2009you} and UCY \cite{lerner2007crowds} datasets, following the leave-one-out strategy \cite{wang2022stepwise,yue2022human,yuan2021agentformer,chib2025lg}. As in prior work \cite{yang2024ia,yu2020spatio,yuan2021agentformer}, we do not use visual information. The model observes 3.2 seconds (8 frames at 2.5 fps) and predicts the next 4.8 seconds. Evaluation employs the standard best-of-$K=20$ protocol \cite{li2019conditional,yue2022human,daniel2021pecnet,yu2020spatio} for fair comparison.

\subsubsection{Metrics.} We use Average Displacement Error (ADE) and Final Displacement Error (FDE) as evaluation metrics. 
% ADE refers to the average displacement error of the predicted trajectory $\widehat{pos}_{\tau+1:T}$ compared to the ground truth trajectory $pos_{\tau+1:T}$, and FDE represents the displacement error of the predicted goal $\widehat{pos}_T$ compared to the ground truth goal $pos_T$.
\begin{equation}
ADE = \frac{1}{T-\tau} \sum_{t=\tau+1}^T \left\| \widehat{pos}_t - pos_t \right\|,
\quad
FDE = \left\| \widehat{pos}_T - pos_T \right\|
\end{equation}
where $\tau$ represents the last time step of the observed trajectory; $T$ represents the last time step of the predicted trajectory; 
% $\widehat{pos}_t$ and $pos_t$ represent the predicted coordinates and ground truth coordinates at time step $t$, respectively. 

Additionally, to validate the effectiveness of the SSOS strategy, we construct the Initial Displacement Error (IDE), which will be discussed in Section~\ref{sec:ablationstudy}.

\subsubsection{Implementation Details.}
Our model builds on the Transformer architecture \cite{vaswani2017attention}. Key configurations: Neighbor Encoder hidden size 128; Seq-CVAE LSTM and latent sizes 256 (see Fig.~\ref{seqCVAE}); neighbor radius 2m. We apply data augmentation (rotation and scaling) and normalize trajectories to start at (0,0). The Interaction Encoder/Generator ($lr=1\times10^{-4}$) and Seq-CVAE ($lr=1\times10^{-3}$, KL weight=5) are trained separately for 500 epochs with Adam and MSE loss. Testing process is autoregressive.
We performed comprehensive data processing and augmentation; the detailed pipeline is provided in our code.

\subsection{Comparison}

\subsubsection{Comparison Baselines.}

We compare our model with representative methods PECNet \cite{daniel2021pecnet} and TF \cite{giuliari2021transformer}, as well as the rencent work \cite{wang2024goal,wang2023trajectory,sang2024mstcnn,lin2025multi}. Additionally, we include some common baselines such as SoPhie \cite{sadeghian2019sophie} and CGNS \cite{li2019conditional}. The baseline results are from officially reported metrics. 

Among these, SoPhie \cite{sadeghian2019sophie}, CGNS \cite{li2019conditional}, STG-DAT \cite{li2021spatio}, and Goal-CurveNet \cite{wang2024goal} leverage visual information, which our approach does not utilize. TF \cite{giuliari2021transformer} is the first to apply Transformers for human trajectory prediction; however, unlike our method, it focuses on individual motion without considering social interactions. 

\vspace{-30pt}

\subsubsection{Evaluation.}
\begin{table*}[h] 
  \centering
  \caption{Results comparison on all datasets. \textbf{Bold}: Best}
  \label{comparison table}
  {\scriptsize
  \begin{tabular}{lccccccccccccc} % 13列（1列方法名 + 1列年份 + 12列数据）
      \toprule % 上边线
      \multirow{2}{*}{Methods} & \multirow{2}{*}{Year} & \multicolumn{2}{c}{ETH} & \multicolumn{2}{c}{Hotel} & \multicolumn{2}{c}{Univ} & \multicolumn{2}{c}{Zara01} & \multicolumn{2}{c}{Zara02} & \multicolumn{2}{c}{Average} \\
      \cmidrule(lr){3-4} \cmidrule(lr){5-6} \cmidrule(lr){7-8} \cmidrule(lr){9-10} \cmidrule(lr){11-12} \cmidrule(lr){13-14}
       &  & ADE & FDE & ADE & FDE & ADE & FDE & ADE & FDE & ADE & FDE & ADE & FDE \\
      \midrule % 中间线
      SoPhie \cite{sadeghian2019sophie} & 2019 & 0.70 & 1.43 & 0.76 & 1.67 & 0.54 & 1.24 & 0.30 & 0.63 & 0.38 & 0.78 & 0.54 & 1.15 \\
      CGNS \cite{li2019conditional} & 2019 & 0.62 & 1.40 & 0.70 & 0.93 & 0.48 & 1.22 & 0.32 & 0.59 & 0.35 & 0.71 & 0.49 & 0.97 \\
      TF \cite{giuliari2021transformer} & 2021 & 0.61 & 1.12 & \textbf{0.18} & 0.30 & 0.35 & 0.65 & 0.22 & \textbf{0.38} & 0.17 & 0.32 & 0.31 & 0.55 \\
      PECNet \cite{daniel2021pecnet} & 2021 & 0.54 & 0.87 & 0.18 & \textbf{0.24} & 0.35 & 0.60 & 0.22 & 0.39 & 0.17 & 0.30 & 0.29 & \textbf{0.48} \\
      SimFuse \cite{habibi2021human}  & 2021 & 0.59 & 1.18 & 0.31 & 0.73 & 0.50 & 1.17 & 0.27 & 0.54 & 0.27 & 0.56 & 0.38 & 0.84 \\
      STG-DAT\cite{li2021spatio}& 2021 & 0.38 & 0.77 & 0.25 & 0.39 & 0.41 & 0.82 & 0.23 & 0.50 & 0.21 & 0.46 & 0.30 & 0.59 \\
      TDAGCN \cite{wang2023trajectory} & 2023 & 0.52 & 0.72 & 0.26 & 0.46 & 0.32 & 0.53 & 0.26 & 0.45 & 0.18 & \textbf{0.29} & 0.31 & 0.49 \\
      MSTCNN \cite{sang2024mstcnn} & 2024 & 0.63 & 0.98 & 0.32 & 0.49 & 0.42 & 0.72 & 0.32 & 0.50 & 0.28 & 0.44 & 0.39 & 0.63 \\
      Goal-CurveNet \cite{wang2024goal} & 2024 & 0.45 & \textbf{0.68} & 0.36 & 0.68 & \textbf{0.31} & \textbf{0.48} & 0.34 & 0.58 & 0.23 & 0.45 & 0.34 & 0.57 \\
      MSWTE-GNN \cite{lin2025multi} & 2025 & 0.51 & 1.04 & 0.23 & 0.44 & 0.32 & 0.64 & 0.24 & 0.45 & 0.23 & 0.42 & 0.31 & 0.60\\ 
      InSyn (Ours) & - & \textbf{0.36} & 0.77 & 0.27 & 0.47 &\textbf{0.31} & 0.54 & \textbf{0.20} & 0.44 & \textbf{0.15} & 0.36 & \textbf{0.26} & 0.52 \\
      \bottomrule % 下边线
  \end{tabular}
  }
\end{table*}

The comparative results are demonstrated in Table~\ref{comparison table}. Compared to the aforementioned methods, our model achieves \textbf{the best average ADE of 0.26}. For goal prediction, the Seq-CVAE achieves an \textbf{average FDE of 0.52}, which is competitive though slightly higher than some state-of-the-art methods \cite{daniel2021pecnet,wang2023trajectory}. This is likely due to its separated design from the main interaction parts. Since ADE measures overall trajectory consistency and better reflects the model's capacity to capture complex pedestrian interactions, we emphasize it as the primary metric for evaluation of interaction modeling.

\textbf{Crowded Scenarios.} In the ETH subset, which represents a bidirectional passageway with frequent pedestrian interactions, InSyn accurately captures complex behaviors such as walking \textit{In Sync} and \textit{Conflict}, and demonstrates strong overall results in this challenging scenario. Similarly, in the UCY dataset, where human-human interactions dominate, InSyn excels on the Zara01 and Zara02 subsets, achieving leading ADE performance. On the Univ subset, our model's performance is comparable to Goal-CurveNet \cite{wang2024goal}, which leverages scene visual information, exhibiting equivalent top-tier results. 

\textbf{Sparse Scenarios.} However, on the Hotel dataset, our model underperforms relatively compared to the original Transformer \cite{giuliari2021transformer}. The Hotel dataset is a smaller scene area, which limits the effectiveness of InSyn's distance-based interaction region partitioning. Moreover, the dataset contains predominantly linear trajectories with sparse interactions. In this scenario, the basic attention mechanism of the original Transformer is sufficient to capture motion patterns effectively, and the additional complex interaction modeling in our model may lead to overfitting, causing excessive reliance on interaction behaviors. This issue is also shared by other complex models such as \cite{wang2023trajectory} and \cite{wang2024goal}, as shown in Table~\ref{comparison table}.

Overall, our model achieves the best average ADE among the compared models and performs particularly well in scenarios with more complex interactions. This demonstrates the model's ability to effectively handle intricate interactions among pedestrians.

\subsection{Ablation Study}
\label{sec:ablationstudy}

\subsubsection{Interaction-Related Components Analysis.} 

For the interaction part, the region partition divides the area centered on the agent into four regions to capture directional interaction effects. The interaction state, proposed in this study, categorizes interactions into three types: \textit{In Sync}, \textit{Conflict}, and \textit{No Neighbor}. To further investigate the roles of these components in the InSyn model, we deconstruct the model into four variations and evaluate their performance, as summarized in Table \ref{RP_IS_table}. ``w/o'' refers to ``without'' a component. The full model (InSyn) includes both RP and IS, while the Baseline model excludes both components. These variations allow us to isolate the individual and combined effects of region partition and interaction state. 

\begin{table}[h] % 使用 table 环境
  \centering
  \caption{ADE Result of Ablation Study on Region Partition (RP) and Interaction State (IS)}
  \label{RP_IS_table}
  {\scriptsize
  \begin{tabular}{lcccccc} % 7列（1列方法名 + 6列数据）
      \toprule % 上边线
      \multirow{2}{*}{Methods} & \multicolumn{6}{c}{ADE} \\ % 合并第二行的2-7列，写上ADE
      \cmidrule(lr){2-7} % 添加横线覆盖所有数据列
        & ETH & Hotel & Univ & Zara01 & Zara02 & Average \\
      \midrule % 中间线
      InSyn    & 0.36 & 0.27 & 0.31 & 0.20 & 0.15 & \textbf{0.26} \\
      \text{w/o}-RP   & 0.61 & 0.33 & 0.37 & 0.28 & 0.23 & 0.36 \\
      \text{w/o}-IS   & 0.46 & 0.27 & 0.36 & 0.25 & 0.29 & 0.33 \\
      Baseline  & 0.60 & 0.28 & 0.41 & 0.29 & 0.24 & 0.36 \\
      \bottomrule % 下边线
  \end{tabular}}
\end{table}

It is evident that the model incorporating both region partition and interaction state, namely the original InSyn, achieves the best performance. This demonstrates that explicitly dividing interaction behaviors enhances the model's ability to extract complex interaction patterns among an agent's neighbors, leading to more reasonable trajectory predictions. 

Specifically, in the absence of region partition, the model may struggle to discern the direction of influence, relying solely on training data distributions to infer future trajectory deviations—a limitation that undermines its generalizability. Similarly, without interaction state, the model would neglect specific interaction relationships between neighboring pedestrians and judge influence solely based on distance and direction. This would underperform when the training and testing sets differ. For instance, if the training set contains more \textit{In Sync} scenarios, the trained model may underestimate repulsive effects in \textit{Conflict} situations. More case studies on the variations are provided in Section~\ref{sec:case study}.

\subsubsection{SSOS Strategy Analysis.} 

Additionally, we conducted an ablation study on the SSOS strategy proposed in this paper. SSOS is designed to mitigate the divergence at the first step when using the Transformer's Encoder-Decoder architecture for numerical time series prediction tasks (detail in Section~\ref{sec:TG}), such as pedestrian trajectory prediction. 
% In this context, steps $0:\tau$ represent observed input to the encoder, while time steps $\tau+1:T$ correspond to the future trajectory to be predicted. We replace the traditional single SOS with the sequence-based SSOS. This approach incorporates the observed values from steps $0:\tau$ as inputs for both the encoder and decoder during the initial stage. During the autoregressive process, the outputs are progressively concatenated and fed back into the Trajectory Generator as inputs. This allows the model to predict the complete trajectory while computing the loss for all outputs (Equation \ref{loss}).
To evaluate the effectiveness of SSOS, we introduce the Initial Displacement Error (IDE):
\begin{equation}
IDE = \left\| \widehat{pos}_{\tau+1} - pos_{\tau+1} \right\|
\end{equation}
where $\tau+1$ denotes the first prediction time step.

The results in Table~\ref{IDE_table} show that the SSOS strategy reduces the average IDE by approximately \textbf{6.58\%} compared to the SOS strategy. This reduction alleviates prediction divergence at time $\tau+1$, enabling smoother transitions between observed and predicted trajectories. Note that the improvement is more significant in Zara datasets, where pedestrians have fewer interactions and rely more on their own historical trajectories. The SSOS strategy, which reinforces the model's encoding of individual motion patterns by integrating the observed sequence, thereby enhances self-expression and leads to greater gains.
\begin{table}[h] % 使用 table 环境
  \centering
  \caption{IDE Result of Ablation Study on SSOS}
  \label{IDE_table}
  {\scriptsize
  \begin{tabular}{lcccccc} % 7列（1列方法名 + 6列数据）
      \toprule % 上边线
      \multirow{2}{*}{Methods} & \multicolumn{6}{c}{IDE} \\ % 合并第二行的2-7列，写上IDE
      \cmidrule(lr){2-7} % 添加横线覆盖所有数据列
      & ETH & Hotel & Univ & Zara01 & Zara02 & Average \\
      \midrule % 中间线
      InSyn-SSOS    & 0.116 & 0.054 & 0.083 & 0.051 & 0.053 & \textbf{0.071} \\
      InSyn-SOS   & 0.118 & 0.051 & 0.085 & 0.062 & 0.065 & 0.076 \\
      \bottomrule % 下边线
  \end{tabular}}
\end{table}
\vspace{-20pt}

\subsection{Case Study}
\label{sec:case study}
\begin{figure*}[h]
    \centering
    \includegraphics[width=4.7in]{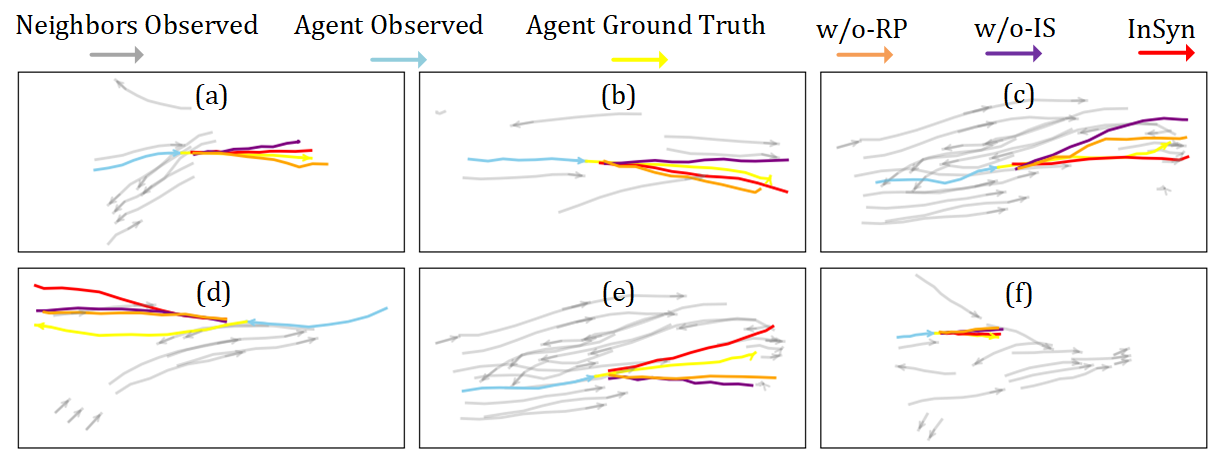} 
    \caption{Case Study of Region Partition and Interaction State. This figure compares three variants of InSyn: (1) without region partition (w/o-RP), (2) without interaction state (w/o-IS), and (3) the full model (InSyn). To isolate the effect of interaction modeling, we \textbf{exclude the Goal Sampler} in this evaluation.}
    \label{fig:case study}
\end{figure*}
To further investigate the effectiveness of region partition and interaction state, we conduct a case study. The cases extracted from the datasets (see Figure \ref{fig:case study}) cover various interaction scenarios, including paired walking, group walking, and pedestrian conflict.

\subsubsection{In Sync Scenarios.} Cases (a), (b), (e), and (f) represent instances where pedestrians walk in coordination with others nearby. In cases (a) and (b), where the target pedestrian moves alongside a companion, InSyn preserves the direction and velocity of the agent's motion at the end of the observation period, closely matching the ground truth. In contrast, both ablated variants produce noticeable deviations. These cases suggest that the influence of the neighboring pedestrian is relatively weak. InSyn effectively models the behaviors by accurately capturing the reduced impact of the neighbor on the agent's motion. Case (e) depicts a group walking situation with multiple neighboring pedestrians moving in sync. Here, only InSyn predicts the correct direction of motion, while both variants mistakenly infer an opposite trajectory, highlighting their failure to capture the low-impact nature of synchronized group behavior. Case (f) presents a simple, slow-moving scenario where all models achieve comparable performance, showing that the base model alone suffices for simple prediction tasks.

\vspace{-10pt}
\subsubsection{Conflict Scenarios.} Cases (c) and (d) illustrate interactions where conflict plays a dominant role. In case (c), although the agent’s last observed motion trends slightly upward, InSyn predicts a flatter, rightward trajectory that better aligns with the ground truth. This adjustment reflects the model's ability to recognize a conflicting pedestrian approaching from above, thereby reducing unnecessary upward movement. The w/o-RP and w/o-IS variants, lacking directional awareness and interaction-type recognition respectively, simply extrapolate the upward motion and thus produce larger errors. Case (d), however, reveals a limitation of InSyn. In this instance, the model appears to overestimate the influence of a pedestrian approaching from below, leading to an excessive upward deviation from the true path. This suggests that while InSyn improves reliability and interpretability in most complex scenarios, further refinement is needed to enhance control precision in handling conflict interactions.

These case studies demonstrate that modeling explicit interaction types and directional influence contributes to more trustworthy and socially-aware predictions. Compared to black-box representations, our approach enhances interpretability, allowing for more transparent reasoning about how different pedestrian behaviors shape trajectory outcomes. This interpretability not only builds trust in the model’s decisions but also provides actionable insights for refining interaction modeling strategies.

\section{Conclusion}

This paper presents InSyn, a trustworthy trajectory prediction model that extends the Transformer architecture to explicitly capture and leverage pedestrian complex interaction patterns. By introducing interpretable structure into otherwise black-box models, InSyn improves both accuracy and interpretability. Experimental results validate the effectiveness of our approach in both dense and sparse scenarios, offering more accurate predictions in complex interaction scenarios. Besides, our proposed SSOS strategy enhances prediction accuracy by mitigating initial-step divergence in numerical sequence modeling. A detailed case study further illustrates how modeling explicit interaction patterns contributes to interpretable and trustworthy predictions in diverse real-world scenarios.
Future work will focus on two key directions. First, we aim to design a more refined approach for interaction identification. Second, we plan to apply the SSOS strategy to other types of numerical sequential data beyond trajectory prediction, exploring its potential in diverse domains.
% \section*{Acknowledgment}
% This work was supported by the Natural Science Foundation of Shanghai (No. 23ZR1422800). The authors would like to thank the SusCom Lab and Dr. Taowen Wang at HKUST(GZ) for their valuable support.
\bibliographystyle{splncs04}
\bibliography{paper}
\end{document}